\documentclass[11pt]{article}
\usepackage{acl2016}
\usepackage{times}
\usepackage{latexsym}

\usepackage{graphicx}
\usepackage{color}
\usepackage{booktabs}
\usepackage{subcaption}
\usepackage{multirow}
\usepackage{amsmath}
\usepackage{amssymb}

\aclfinalcopy 

\DeclareMathOperator \real{\mathbb{R}}

\DeclareMathOperator*{\sig}{sig}

\newcommand{\Ni}{({\em i})~}
\newcommand{\Nii}{({\em ii})~}
\newcommand{\Niii}{({\em iii})~}

\title{Machine Translation Evaluation Meets Community Question Answering}

\author{Francisco Guzm\'an\hbox{\rm ,} Llu\'is M\`arquez \and Preslav Nakov\\
        Arabic Language Technologies Research Group\\ 
        Qatar Computing Research Institute, HBKU\\ 
        {\tt \{fguzman,lmarquez,pnakov\}@qf.org.qa}}
         
\date{}

\begin{document}

\maketitle

\begin{abstract}
We explore the applicability of machine translation evaluation (MTE) methods to a very different problem: answer ranking in community Question Answering. In particular, we adopt a pairwise neural network (NN) architecture, which incorporates MTE features, as well as rich syntactic and semantic embeddings, and which efficiently models complex non-linear interactions. The evaluation results show state-of-the-art performance, with sizeable contribution from both the MTE features and from the pairwise NN architecture.
\end{abstract}

\vspace{2pt}
\section{Introduction and Motivation}

\begin{figure*}[ht]
\centering
\includegraphics [width=\textwidth]{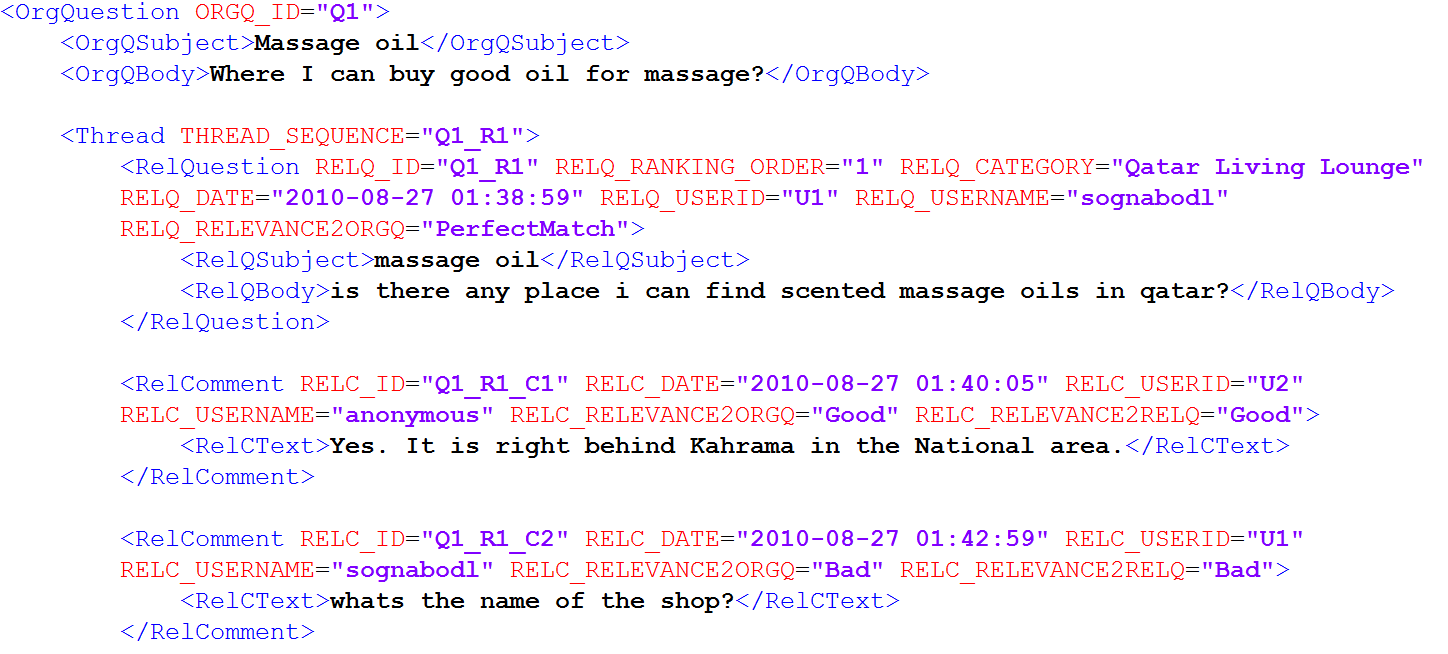}
\caption{Annotated English question from the CQA-QL corpus. Shown are the first two comments only.}
\label{fig:EnglishXMLFig}
\end{figure*}

In a community Question Answering (cQA) task, we are given a question from a community forum and a thread of associated text comments intended to answer the given question; and the goal is to rank the comments according to their appropriateness to the question. Since cQA forum threads are noisy (e.g., because over time people tend to engage in discussion and to deviate from the original question), as many comments are not answers to the question, the challenge lies in learning to rank all \emph{good} comments above all \emph{bad} ones.

Here, we adopt the definition and the datasets from SemEval--2016 Task 3 \cite{nakov-EtAl:2016:SemEval} on ``Community Question Answering'', 
focusing on subtask A (Question-Comment Similarity) only.\footnote{SemEval-2016 Task 3 had two more subtasks: subtask B on Question-Question Similarity, and subtask C on Question-External Comment Similarity, which are out of our scope. However, they could be potentially addressed within our general MTE-NN framework, with minor variations.} See the task description paper and the task website\footnote{\texttt{http://alt.qcri.org/semeval2016/task3/}} for more detail. An annotated example is shown in Figure~\ref{fig:EnglishXMLFig}.

In this paper, we tackle the task from a novel perspective: by using ideas from machine translation evaluation (MTE) to decide on the \emph{quality} of a comment. 
In particular, we extend our MTE neural network framework from~\cite{guzman-EtAl:2015:ACL-IJCNLP}, showing that it is applicable to the cQA task as well. We believe that this neural network is interesting for the cQA problem because: \Ni it works in a pairwise fashion, i.e., given two translation hypotheses and a reference translation to compare to, the network decides which translation hypothesis is better, which is appropriate for a ranking problem; \Nii it allows for an easy incorporation of rich syntactic and semantic embedded representations of the input texts, and it efficiently models complex non-linear relationships between them; \Niii it uses a number of machine translation evaluation measures that have not been explored for the cQA task before, e.g., \textsc{Ter} \cite{Snover06astudy}, \textsc{Meteor} \cite{Lavie:2009:MMA}, and \textsc{Bleu} \cite{Papineni:Roukos:Ward:Zhu:2002}.

The analogy we apply to adapt the neural MTE architecture to the cQA problem is the following: given two comments $c_1$ and $c_2$ from the question thread---which play the role of the two competing 
translation hypotheses---we have to decide whether $c_1$ is a better answer than $c_2$ to question $q$---which plays the role of the translation reference.
If we have a function $f(q,c_1,c_2)$ to make this decision, then we can rank the finite list of comments in the thread by comparing all possible pairs and by accumulating for each comment the scores for it given by $f$.

From a general point of view, 
MTE and the cQA task addressed in this paper seem similar: both reason about the similarity of two competing texts against a reference text in order to decide which one is better. However, there are some profound differences, which have implications on how each task is solved. 

In MTE, the goal is to decide whether a hypothesis translation conveys the same meaning as the reference translation. In cQA, it is to determine whether the comment is an appropriate answer to the question. Furthermore, in MTE we can expect shorter texts, which are typically much more similar. 
In contrast, in cQA, the question and the intended answers might differ significantly both in terms of length and in lexical content. Thus, it is not clear a priori whether the MTE network can work well to address the cQA problem. Here, we show that the analogy is not only convenient, but also that using it can yield state-of-the-art results for the cQA task.

To validate our intuition, we present series of experiments using the publicly available SemEval-2016 Task 3 datasets, with focus on subtask A. We show that a na\"{i}ve application of the MTE architecture and features on the cQA task already yields results that are largely above the task baselines. Furthermore, by adapting the models with in-domain data, and adding lightweight task-specific features, we are able to boost our system to reach state-of-the-art performance. 

More interestingly, we analyze the contribution of several features and parts of the NN architecture by performing an ablation study. We observe that every single piece contributes important information to achieve the final performance. While task-specific features are crucial, other aspects of the framework are relevant as well: syntactic embeddings, machine translation evaluation measures, and pairwise training of the network.

The rest of the paper is organized as follows:
Section~\ref{sec:related} introduces some related work.
Section~\ref{sec:model} presents the overall architecture of our MTE-inspired NN framework for cQA.
Section~\ref{sec:features} summarizes the features we use in our experiments.
Section~\ref{sec:exps} describes the experimenal settings and presents the results.
Finally, Section~\ref{sec:conclusion} offers further discussion
and presents the main conclusions.

\section{Related Work} 
\label{sec:related}

Recently, many neural network (NN) models have been applied to cQA tasks: e.g., \emph{question-question similarity} \cite{zhou-EtAl:2015,dossantos-EtAl:2015,LeiJBJTMM16}
and \emph{answer selection} \cite{severyn2015sigir,wang-nyberg:2015:ACL-IJCNLP,shen2015word,feng2015applying,tan2015lstm}.
Most of these papers concentrate on providing advanced neural architectures in order to better model the problem at hand. 
However, our goal here is different: we extend and reuse an existing pairwise NN framework from a different but related problem.

There is also work that uses machine translation models as a features
for cQA~\cite{Berger:2000:BLC:345508.345576,Echihabi:2003:NAQ:1075096.1075099,jeon2005finding,Soricut:2006:AQA:1127331.1127342,riezler-EtAl:2007:ACLMain,li2011improving,Surdeanu:2011:LRA:2000517.2000520,tran-EtAl:2015:SemEval} 
e.g., a variation of IBM model 1, to compute the probability that the question is a possible ``translation'' of the candidate answer.
Unlike that work, here we port an entire MTE framework to the cQA problem.
A preliminary version of this work was presented in \cite{SemEval2016:task3:MTE-NN}.

\label{sec:intro}

\section{Neural Model for Answer Ranking}
\label{sec:model}

\begin{figure}[t]
\centering
\hspace*{-2mm}\includegraphics[width=.5\textwidth]{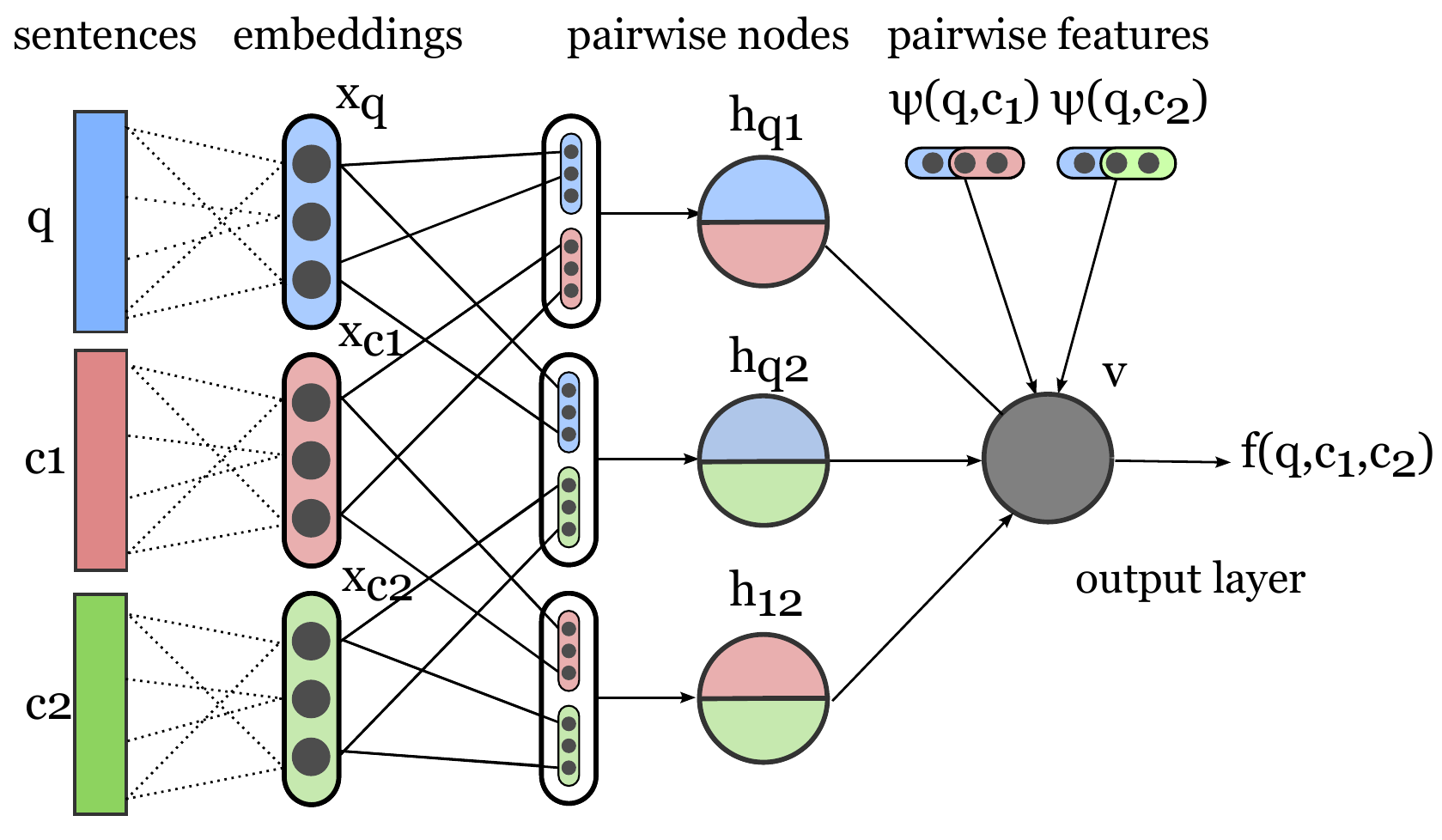}
\caption{\label{fig:architecture} Overall architecture of the NN.}
\end{figure}

The NN model we use for answer ranking is depicted in Figure~\ref{fig:architecture}. It is a direct adaptation of our feed-forward NN for MTE \cite{guzman-EtAl:2015:ACL-IJCNLP}. 
%
Technically, we have a binary classification task with input $(q,c_1,c_2)$, which should output 1 if $c_1$ is a better answer to $q$ than $c_2$, and 0 otherwise.
The network computes a sigmoid function $f(q,c_1,c_2)=\sig(\mathbf{w^T_v}\phi(q,c_1,c_2) + b_v)$, 
where $\phi(x)$  transforms the input $x$ through the hidden layer, $\mathbf{w_v}$ are the weights from the hidden layer to the output layer, and $b_v$ is a bias term. 

We first map the question and the comments to a fixed-length vector $\left[ \mathbf{x}_q, \mathbf{x}_{c_1},  \mathbf{x}_{c_2} \right]$ using syntactic and semantic embeddings. Then, we feed this vector as input to the neural network, which models three types of interactions, using different groups of nodes in the hidden layer. 
There are two \emph{evaluation} groups $\mathbf{h_{q1}}$ and $\mathbf{h_{q2}}$
that model how good each comment $c_i$ is to the question $q$.
The input to these groups are the concatenations $\left[ \mathbf{x}_q, \mathbf{x}_{c_1} \right]$ and $\left[\mathbf{x}_{q}, \mathbf{x}_{c_2} \right]$, respectively.
The third group of hidden nodes $\mathbf{h_{12}}$, which we call \emph{similarity} group, models how close $c_1$ and $c_2$ are. Its input is $\left[ \mathbf{x}_{c_1}, \mathbf{x}_{c_2} \right]$.
This might be useful as highly similar comments are likely to be comparable in appropriateness, irrespective of whether they are good or bad answers in absolute terms.

In summary, the transformation $\phi(q,c_1,c_2) = [\mathbf{h_{q1}}, \mathbf{h_{q2}}, \mathbf{h_{12}}]$ can be written as
\vspace*{-0.5mm}
\begin{eqnarray*}
 \mathbf{h_{qi}} &=& g( \mathbf{W_{qi}} \mathbf{\left[ \mathbf{x}_q, \mathbf{x}_{c_i} \right]} + \mathbf{b_{qi}}),\, i=1,2\\
 \mathbf{h_{12}} &=& g( \mathbf{W_{12}} \mathbf{\left[ \mathbf{x}_{c_1}, \mathbf{x}_{c_2} \right]} +  \mathbf{b_{12}}),
\vspace*{-1mm}
\end{eqnarray*}
\noindent where $g(.)$ is a non-linear activation function (applied component-wise), $\mathbf{W} \in \real^{H \times N}$ are the associated weights between the input layer and the hidden layer, and $\mathbf{b}$ are the corresponding bias terms.

We use $\tanh$ as an activation function, rather than $\sig$, to be consistent with how the word embedding vectors we use were generated.

The model further allows to incorporate external sources of information in the form of \emph{skip arcs} that go directly from the input to the output, skipping the hidden layer. These arcs represent pairwise \emph{similarity} feature vectors between $q$ and either $c_1$ or $c_2$. In these feature vectors, we encode MT evaluation measures (e.g., \textsc{Ter}, \textsc{Meteor}, and \textsc{Bleu}), cQA task-specific features, etc. See Section~\ref{sec:features} for detail about the features implemented as skip arcs. In the figure, we indicate these pairwise external feature sets as $\psi(q,c_1)$ and  $\psi(q,c_2)$. When including the external features, the activation at the output is
$f(q,c_1,c_2)=\sig(\mathbf{w^T_v} [\phi(q,c_1,c_2), \psi(q,c_1), \psi(q,c_2)] + b_v)$.


\section{Features}
\label{sec:features}
We experiment with three kinds of features:
(\emph{i})~input embeddings,
(\emph{ii})~features from MTE \cite{guzman-EtAl:2015:ACL-IJCNLP}
and
(\emph{iii})~task-specific features from SemEval-2015 Task 3 \cite{nicosia-EtAl:2015:SemEval}.

\paragraph{A. Embedding Features}
We used two types of vector-based embeddings to encode the input texts $q$, $c_1$ and $c_2$:
(1)~{\bf \textsc{Google\_vectors}}: 
300-dimensional embedding vectors,
trained on 100 billion words from Google News \cite{mikolov-yih-zweig:2013:NAACL-HLT}.
The encoding of the full text is just the average of the word embeddings.
(2)~{\bf \textsc{Syntax}}: We parse the entire question/comment using the Stanford neural parser \cite{socher-EtAl:2013:ACL2013}, and we use the final 25-dimensional vector that is produced internally as a by-product of parsing.


Also, we compute cosine similarity features with the above vectors:
$\cos(q,c_1)$ and $\cos(q,c_2)$.



\paragraph{B. MTE features}
We use the following  
MTE metrics ({\bf \textsc{MTfeats}}),
which compare the similarity between the question and a candidate answer:
(1)~\textsc{Bleu} \cite{Papineni:Roukos:Ward:Zhu:2002};
(2)~\textsc{NIST} \cite{Doddington:2002:AEM};
(3)~\textsc{TER} v0.7.25 \cite{Snover06astudy}.
(4)~\textsc{Meteor} v1.4 \cite{Lavie:2009:MMA} with paraphrases;
(5)~Unigram ~\textsc{Precision}; 
(6)~Unigram ~\textsc{Recall}.

%


{\bf \textsc{BLEUcomp}}. We 
further use as features various components 
involved in the computation of \textsc{Bleu}:
$n$-gram precisions,
$n$-gram matches,
total number of $n$-grams ($n$=1,2,3,4),
lengths of the hypotheses and of the reference, 
length ratio between them,
and \textsc{Bleu}'s brevity penalty.

\paragraph{C. Task-specific features}
First, we train domain-specific vectors using \textsc{word2vec}
on all available QatarLiving data, both annotated and raw ({\bf \textsc{QL\_vectors}}).

Second, we compute various easy task-specific features ({\bf \textsc{Task\_features}}),
most of them proposed for the 2015 edition of the task \cite{nicosia-EtAl:2015:SemEval}.
This includes some comment-specific features:
(1)~number of URLs/images/emails/phone numbers;
(2)~number of occurrences of the string ``thank'';\footnote{When an author thanks somebody, this post is typically a bad answer to the original question.}
(3)~number of tokens/sentences;
(4)~average number of tokens;
(5)~type/token ratio;
(6)~number of nouns/verbs/adjectives/adverbs/ pronouns;
(7)~number of positive/negative smileys;
(8)~number of single/double/triple exclamation/interrogation symbols;
(9)~number of interrogative sentences (based on parsing);
(10)~number of words that are not in \textsc{word2vec}'s Google News vocabulary.\footnote{Can detect slang, foreign language, etc., which would indicate a bad answer.}
Also some question-comment pair features:
(1)~question to comment count ratio in terms of sen\-tences/tokens/nouns/verbs/adjectives/adverbs/pro\-nouns;
(2)~question to comment count ratio of words that are not in \textsc{word2vec}'s Google News vocabulary.
Finally, we also have two meta features:
(1)~is the person answering the question the one who asked it;
(2)~reciprocal rank of the comment in the thread.


\section{Experiments and Results}
\label{sec:exps}
  
\label{subsec:setting}
We experiment with the data from SemEval-2016 Task 3.
The task offers a higher quality training dataset \textsc{train-part1},
which includes 1,412 questions and 14,110 answers,
and a lower-quality \textsc{train-part2} with 382 questions and 3,790 answers.
We train our model on \textsc{train-part1} with hidden layers of size 3
for 100 epochs with minibatches of size 30, regularization of 0.005,
and a decay of 0.0001, using stochastic gradient descent with adagrad~\cite{Duchi11};
we use Theano \cite{bergstra+al:2010-scipy} for learning.
We normalize the input feature values to the $[-1; 1]$ interval using minmax,
and we initialize the network weights by sampling from a uniform distribution as in \cite{Xavier10}.
We train the model using all pairs of good vs. bad comments,
in both orders, ignoring ties.

At test time, we get the full ranking by scoring all possible pairs,
and we accumulate the scores at the comment level. 

We evaluate the model on \textsc{train-part2} after each epoch,
and ultimately we keep the model that achieves the highest accuracy;\footnote{We also tried Kendall's Tau ($\tau$), but it performed worse.}
in case of a tie, we prefer the parameters from an earlier epoch.
We selected the above parameter values on the \textsc{dev} dataset (244 questions and 2,440 answers) using the full model,
and we used them for all experiments below,
where we evaluate on the official \textsc{test} dataset (329 questions and 3,270 answers).
We report mean average precision (MAP),
which is the official evaluation measure,
and also 
average recall (AvgRec) and mean reciprocal rank (MRR).

\label{subsec:results}
\vspace{3pt}
\subsection{Results}
\begin{table}
\centering
\small
\begin{tabular}{lccc}
\toprule
{\bf System} & {\bf MAP} & {\bf AvgRec} & {\bf MRR}\\
\midrule
MTE-CQA$_{pairwise}$ & {\bf 78.20} & {\bf 88.01} & {\bf 86.93} \\
MTE-CQA$_{classification}$ & 77.62 & 87.85 & 85.79 \\ 
MTE$_{vanilla}$ & 70.17 & 81.84 & 78.60 \\
Baseline$_{time}$   & 59.53 & 72.60 & 67.83 \\
Baseline$_{rand}$ & 52.80 & 66.52 & 58.71 \\
\bottomrule
\end{tabular}
\caption{Main results on the ranking task.}\label{tab:mainresults}
\vspace{-5pt}
\end{table}


Table~\ref{tab:mainresults} shows the evaluation results
for three configurations of our MTE-based cQA system.
We can see that the vanilla MTE system (MTE$_{vanilla}$), 
which only uses features from our original MTE model, i.e., it does not have any task-specific features (\textsc{Task\_Features} and \textsc{QL\_vectors}), performs surprisingly well 
despite the differences in the MTE and cQA tasks.
It outperforms a random baseline (Baseline$_{rand}$)
and a chronological baseline that assumes that early comments are 
better than later ones (Baseline$_{time}$) by large margins: by about 11 and 
17 MAP points absolute, respectively. For the other two measures the results are similar. 

We can further see that adding the task-specific features in MTE-CQA$_{pairwise}$
improves the results by another 8 MAP points absolute.
Finally, the second line shows that adapting the network to do 
classification (MTE-CQA$_{classification}$), giving it a question and a single 
comment as input, yields a performance drop of 0.6 MAP points absolute
compared to the proposed pairwise learning model. 
Thus, the pairwise training strategy is confirmed to be better for the ranking task, 
although not by a large margin. 
 
\begin{table}[tbh]
\centering
\small
\begin{tabular}{lcccc}
\toprule
{\bf System} & {\bf MAP} & {\bf AvgRec} & {\bf MRR} & {\bf $\Delta_{\hbox{\scriptsize MAP}}$ }\\
\midrule
MTE-CQA   & {\bf 78.20} & {\bf 88.01} & {\bf 86.93} & \\
\midrule
$-$\textsc{BLEUcomp} & 77.83 & 87.85 & 86.32 & -0.37\\
$-$\textsc{MTfeats}  & 77.75 & 87.76 & 86.01 & -0.45\\
$-$\textsc{Syntax}   & 77.65 & 87.65 & 85.85 & -0.55\\
$-$\textsc{Google\_vect.} & 76.96 & 87.66 & 84.72 & -1.24\\
$-$\textsc{QL\_vectors}  & 75.83 & 86.57 & 83.90 & -2.37\\
$-$\textsc{Task\_Feats.} & 72.91 & 84.06 & 78.73 & -5.29\\
\bottomrule
\end{tabular}
\caption{Results of the ablation study.}\label{tab:ablation}
\end{table}

Table~\ref{tab:ablation} presents the results of an ablation study,
where we analyze the contribution of various features and feature groups 
to the performance of the overall system. For the purpose, we study 
$\Delta_{\hbox{\scriptsize MAP}}$, i.e., the absolute drop in MAP when 
the feature group is excluded from the full system.

Not surprisingly, the most important turn out to be the \textsc{Task\_Features} 
(contributing over five MAP points) as they handle important information sources 
that are not available to the system from other feature groups, e.g., the 
reciprocal rank alone contributes about two points.

Next in terms of importance come word embeddings, \textsc{QL\_vectors} 
(contributing over 2 MAP points), trained on text from the target forum, 
QatarLiving. Then come the \textsc{Google\_vectors} (contributing over one 
MAP point), which are trained on 100 billion words, and thus are still useful 
even in the presence of the domain-specific \textsc{QL\_vectors}, which are in 
turn trained on four orders of magnitude less data.

Interestingly, the MTE-motivated \textsc{Syntax} vectors contribute half a MAP 
point, which shows the importance of modeling syntax for this task.
The other two MTE features, \textsc{MTfeats} and \textsc{BLEUcomp}, together 
contribute 0.8 MAP points. It is interesting that the \textsc{Bleu} components 
manage to contribute on top of the \textsc{MTfeats}, which already contain several 
state-of-the-art MTE measures, including \textsc{Bleu} itself.
This is probably because the other features we have do not model $n$-gram 
matches directly.

\begin{table}[t]
\centering
\small
\begin{tabular}{lccc}
\toprule
{\bf System} & {\bf MAP} & {\bf AvgRec} & {\bf MRR}\\
\midrule
1st~{\scriptsize \cite{SemEval2016:task3:KeLP}} & 79.19 & 88.82 & 86.42\\
\textbf{\em MTE-CQA} & \textbf{\em 78.20} & \textbf{\em 88.01} & \textbf{\em 86.93}\\
2nd~{\scriptsize \cite{SemEval2016:task3:ConvKN}}  & 77.66 & 88.05 & 84.93\\
3rd~{\scriptsize \cite{SemEval2016:task3:SemanticZ}} & 77.58 & 88.14 & 85.21\\\
$\ldots$ & $\ldots$ & $\ldots$ & $\ldots$\\
Average & 73.54 & 84.61 & 81.54\\
$\ldots$ & $\ldots$ & $\ldots$ & $\ldots$\\
12th (Worst) & 62.24 & 75.41 & 70.58\\
\bottomrule
\end{tabular}
\caption{Comparative results with the best SemEval-2016 Task 3, subtask A systems.}\label{tab:comparison}
\vspace{-3pt}

\end{table}

Finally, Table~\ref{tab:comparison} puts the results in perspective.
We can see that our system MTE-CQA would rank first on MRR, second on MAP, 
and fourth on AvgRec in SemEval-2016 Task 3 competition.\footnote{The full results 
can be found on the task website: http://alt.qcri.org/semeval2016/task3/index.php?id=results}
These results are also 5 and 16 points above the average and the worst systems, respectively.
\\\vspace{5pt}

This is remarkable given the lightweight task-specific features we use,
and confirms the validity of the proposed neural approach to produce state-of-the-art systems
for this particular cQA task.
%

\section{Conclusion}
\label{sec:conclusion}
We have explored the applicability of machine translation evaluation methods
to answer ranking in community Question Answering, a seemingly very different task,
where the goal is to rank the comments in a question-answer thread according to their appropriateness to the question, placing all good comments above all bad ones.

In particular, 
we have adopted a pairwise neural network architecture, which incorporates MTE features, as well as rich syntactic and semantic embeddings of the input texts that are non-linearly combined in the hidden layer.
The evaluation results on benchmark datasets have shown state-of-the-art performance, with sizeable contribution from both the MTE features and from the network architecture. This is an interesting and encouraging result, as given the difference in the tasks, it was not a-priori clear that an MTE approach would work well for cQA.

In future work, we plan to incorporate other similarity measures 
and better task-specific features into the model.
We further want to explore the application of this architecture to other semantic similarity 
problems such as question-question similarity, and textual entailment.

%

\section*{Acknowledgments}
This research was performed by the Arabic Language Technologies (ALT) group at the Qatar Computing Research Institute (QCRI), Hamad bin Khalifa University, part of Qatar Foundation. It is part of the Interactive sYstems for Answer Search (Iyas) project, which is developed in collaboration with MIT-CSAIL.

\bibliography{acl2016}
\bibliographystyle{acl2016}

\end{document}